%% file: main.tex
\documentclass[lettersize,journal]{IEEEtran}
\def\endthebibliography{%
	\def\@noitemerr{\@latex@warning{Empty `thebibliography' environment}}%
	\endlist
}
\usepackage{amsmath,amsfonts}
\usepackage{algorithmic}
\usepackage{algorithm}
\usepackage{array}
\usepackage[caption=false,font=normalsize,labelfont=sf,textfont=sf]{subfig}
\usepackage{textcomp}
\usepackage{stfloats}
\usepackage{url}
\usepackage{verbatim}
\usepackage{graphicx}
\usepackage{multirow}
\usepackage{cite}
\usepackage{tabularx}
\usepackage{booktabs}
\usepackage{algorithm}
\usepackage{algorithmic}
\usepackage{booktabs}  
\usepackage{graphicx}
\usepackage{bbding}
\usepackage{multirow, multicol}
\usepackage{colortbl}
\begin{document}

\title{PITN: Physics-Informed Temporal Networks \\for Cuffless Blood Pressure Estimation}

\author{Rui~Wang,
        Mengshi Qi,~\IEEEmembership{Member,~IEEE,}
        Yingxia Shao,
        Anfu Zhou,
        Huadong Ma,~\IEEEmembership{Fellow,~IEEE} 
\thanks{This work is partly supported by the Funds for the NSFC Project under Grant 62202063, Beijing Natural Science Foundation (L243027 and L223002), the Innovation Research Group Project of the NSFC under Grant 61921003. (\emph{Corresponding author: Mengshi Qi~(email:~qms@bupt.edu.cn)})}
\thanks{R. Wang, M. Qi, Y. Shao, A. Zhou, and H. Ma are with State Key Laboratory of Networking and Switching Technology, Beijing University of Posts and Telecommunications, China (e-mail: \{wr, qms, shaoyx, zhouanfu, mhd\}@bupt.edu.cn).}
}

\markboth{}%
{Shell \MakeLowercase{\textit{et al.}}: A Sample Article Using IEEEtran.cls for IEEE Journals}

\maketitle

\begin{abstract}
Monitoring blood pressure with non-invasive sensors has gained popularity for providing comfortable user experiences, one of which is a significant function of smart wearables. Although providing a comfortable user experience, such methods are suffering from the demand for a significant amount of realistic data to train an individual model for each subject, especially considering the invasive or obtrusive BP ground-truth measurements. To tackle this challenge, we introduce a novel physics-informed temporal network~(PITN) with adversarial contrastive learning to enable precise BP estimation with very limited data. Specifically, we first enhance the physics-informed neural network~(PINN) with the temporal block for investigating BP dynamics' multi-periodicity for personal cardiovascular cycle modeling and temporal variation. We then employ adversarial training to generate extra physiological time series data, improving PITN's robustness in the face of sparse subject-specific training data. Furthermore, we utilize contrastive learning to capture the discriminative variations of cardiovascular physiologic phenomena. This approach aggregates physiological signals with similar blood pressure values in latent space while separating clusters of samples with dissimilar blood pressure values. Experiments on three widely-adopted datasets with different modailties (\emph{i.e.,} bioimpedance, PPG, millimeter-wave) demonstrate the superiority and effectiveness of the proposed methods over previous state-of-the-art approaches. The code is available at~\url{https://github.com/Zest86/ACL-PITN}. 
\end{abstract}


\section{Introduction}
Continous monitoring of vital physiological signs such as heart rate (HR), blood pressure (BP), and respiration rate (RR), is essential and has become popular on smart wearables. The blood circulatory system can be considered as a whole-body interconnecting organ, hence, blood pressure~(BP) is one of the most crucial indicators of cardiovascular health~\cite{kireev2022continuous}. Two common irregularities in BP are hypertension~(high blood pressure) and hypotension~(low blood pressure) which can easily be overlooked despite indicating certain cardiovascular diseases such as stroke or chronic kidney issues. Therefore, accurate blood pressure estimation holds significant importance in monitoring everyday human health. To enable continuous monitoring of BP values, various wearable devices have been integrated~\cite{zhao_emerging_2023} to mine valuable information for health analysis, such as graphene electronic tattoos and multiple sensors. These devices enable precise BP estimation using diverse multimodal signals, including bioimpedance, millimeter-wave, and photoplethysmography (PPG). This casts a problem of how these multi-sensor data should be analyzed to predict the subject's health information. They significantly enhance user experience and contribute to the advancement of digital health.

\begin{figure}[t]
	\includegraphics[width=0.47\textwidth]{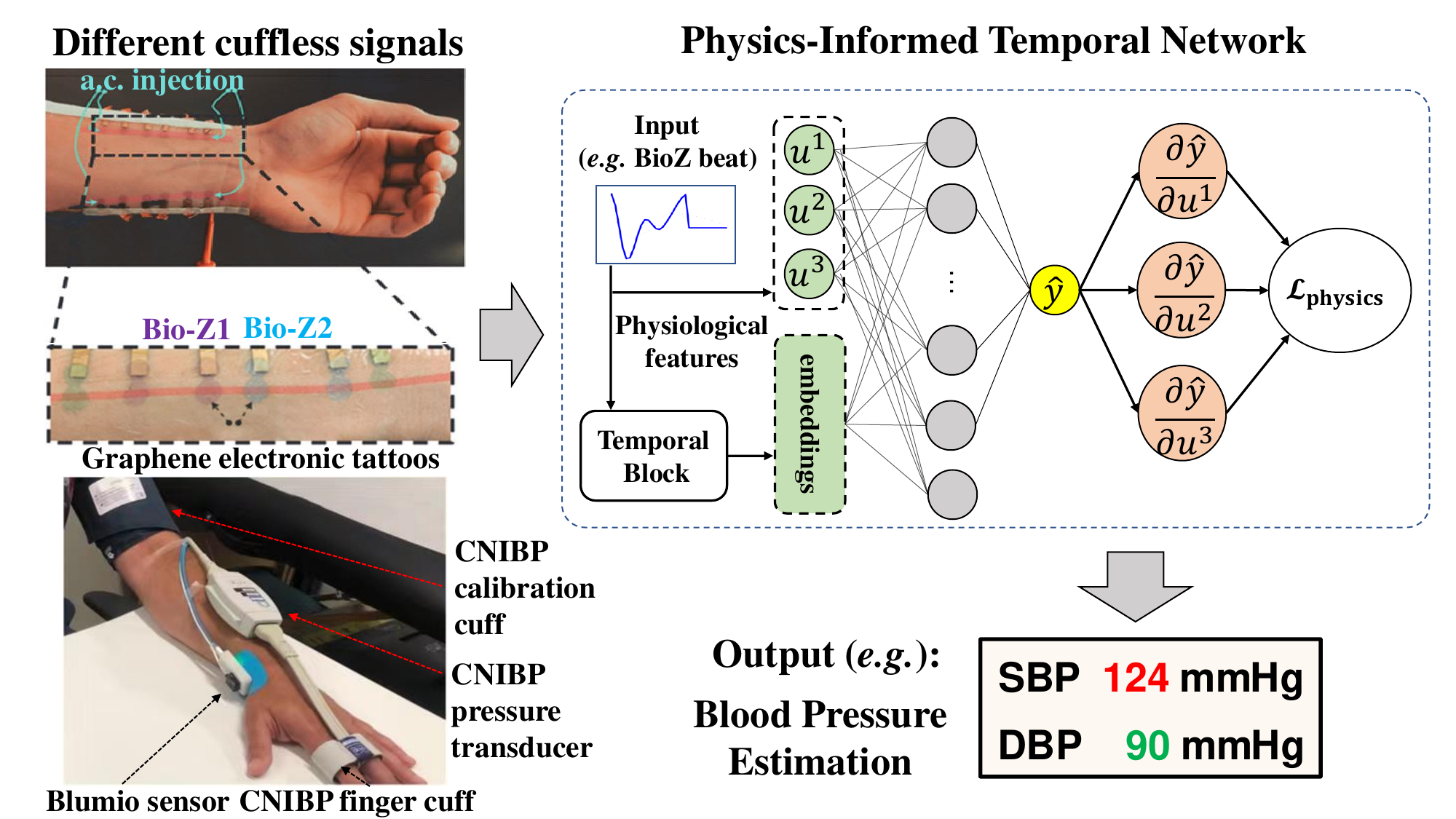} 
	\centering
	\caption{Illustration of the cuffless blood pressure estimation task by inputting bioimpedance signals in graphene-HGCPT dataset (left top)~\cite{kireev2022continuous}, PPG and millimeter-wave signal in blumio dataset (left bottom)~\cite{blumio}.}
	\label{fig: task illustration}
        
\end{figure}

Traditionally, blood pressure is measured using a cuff to compress the arm, which causes an unpleasant user experience~\cite{gonzalez2023benchmark}. 
As a result, cuffless BP estimation has gained significant development in recent years to improve user comfort. A lot of ambulatory BP monitoring platforms provide multiple solutions for continuous cuffless BP estimation in non-clinical (ambulatory) settings using wearable devices like bioimpedance~\cite{sel2023continuous, sel2023physics, ibrahim_cuffless_2022} or millimeter wave sensor~\cite{blumio}.
Considering that these electrical signals are frequently a multi-variable measurement utilized to predict physiological states, recent developments in AI algorithms, especially deep learning models, can provide great opportunities for cuffless BP estimation. This involves inferring BP using data from cuffless wearable devices (e.g., Blumio sensor, graphene electronic tattoos), by modeling the complex inherent input-output temporal relationships in physiological systems. However, these methods usually require collecting significant amounts of training data as well as ground truth labels for each subject to train a powerful model. In fact, invasive or obtrusive medical-grade measurement systems (\emph{e.g.}, arterial lines or cuff-based (auscultatory or oscillometric) sphygmomanometers for peripheral blood pressure monitoring), are cumbersome and impractical in certain contexts. To address these challenges, there is a pressing need to develop advanced models for time series data that can achieve accurate BP estimation with reduced reliance on ground truth data.

So far, extensive efforts~\cite{sel2023continuous,cao2021crisp,liang2023airbp, zhang2019real,sel2023physics} have been devoted to cuffless BP estimation, which utilizes various methods for modeling, such as Finite Element Analysis~(FEA)~\cite{sel2023continuous}, Bidirectional Long Short Term Memory (BiLSTM)-based Hybrid Neural Network (HNN)~\cite{cao2021crisp} and Autoencoder~\cite{liang2023airbp}, etc. These models work in a pure data-driven framework, without utilizing domain knowledge in biological systems, which makes them effective in numerous training data situations but fail in personalized modeling. In this paper, we focus on improving the personal modeling ability of previous beat-to-beat BP estimation models and data augmentation methods which provide more training examples. Furthermore, the emerging physics-informed neural networks (PINNs)~\cite{raissi2019physics} which leverage underlying physics laws structured by generalized partial differential equations (PDEs), have proven their superiority in related fields such as structural mechanics~\cite{kapoor_physics-informed_2024}, mechanical systems~\cite{xu_physics-constraint_2024}, and biomedical informatics~\cite{oszkinat_uncertainty_2023}. Compared with the above-mentioned pure data-driven BP estimation approaches, PINN utilizes physiological prior to formalizing additional physics residual for constraint on model training, which enhances BP estimation with reduced labels for training~\cite{sel2023physics}.  

Vanilla PINNs take coordinates as input, which ignores the potential temporal relation between input time series data~\cite{zhao2023pinnsformer}. However, beat-to-beat BP estimation is a continuous time series data mining task, where input time series waveform contains rich temporal information.  Furthermore, there are some Transformer-based methods that utilize the long-term modeling ability for cuffless blood pressure estimation~\cite{ma_kd-informer_2023}. Specifically, Transformer-based models adopt the attention mechanism or its variants to capture the pair-wise temporal dependencies among time points~\cite{zhang2023warpformer}. However, it is still challenging for the attention mechanism to directly identify reliable dependencies from scattered time points, as these temporal dependencies can be deeply obscured by intricate temporal patterns~\cite{yu2023mpre}. To bridge the gap between PINN and temporal modeling, we propose a novel Physics-Informed Temporal Network (PITN) for cuffless blood pressure estimation, which enhances PINN's ability in temporal data mining through a novel temporal block for personal cardiovascular cycle modeling as shown in Fig.~\ref{fig: task illustration}. The proposed temporal block enables temporal variation modeling through transformation of 1D time series to 2D space and is suitable for personalized cardiovascular cycle modeling.

For data scarcity, one promising way is adversarial training, which is used to improve the model's robustness. Several recent works show that adversarial examples can be used to improve model performance in computer vision (CV)~\cite{aug-nerf}, natural language processing~\cite{zhang2022adversarial}, and other domains. 
These previous works make it promising to use adversarial samples for data augmentation in time series analysis. Unfortunately, limited work adopts such a similar adversarial training strategy in medical time series data, which is our main focus of data augmentation methods in this work.

In addition, contrastive representation learning has been shown as a promising manner to mitigate data scarcity~\cite{liu2021self}. However, extending contrastive learning paradigms to time series presents significant challenges, especially in the health domain with unique characteristics (\emph{e.g.}, low frequency, and high sensitivity~\cite{yu2018artificial}). Although, some previous work adopt self-supervised
contrastive learning to mitigate the challenge of label scarcity in medical
time series~\cite{kiyasseh2021clocs, liu2021self}.
These self-supervised methods require a large amount of data for pre-training, which is an emergent weakness. Besides, Sel~\emph{et al.}~\cite{sel2023physics} point out that for deep learning-based BP estimation models, as the training data increases, personalized correlation fluctuates.

For this part, to the best of our knowledge, there is no related work focusing on improving this insufficiency. In this work, we utilize contrastive learning to address this problem by adding soft constrain on latent embedding, to enable the deep model to learn from physiological signals with similar BP values. Through using generated adversarial examples, our proposed method can offer extra alternatives for the deep model to capture the discriminative variations of cardiovascular physiologic phenomena.

The main advantages of our proposed framework are three-fold. First, PITN extends vanilla PINN models's ability in time series modeling by capturing intraperiodic temporal features. Second, we further enhance the PITN by incorporating adversarial training which generates extra adversarial samples to ensure sufficient robustness. These perturbed samples also serve as data augmentation to boost model training. Last but not least, the proposed contrastive learning leverages adversarial samples to better capture the discriminative BP dynamics. Our contributions can be summarized as follows:
\begin{itemize}
	\item We propose an end-to-end Physics-Informed Temporal Network (PITN) for cuffless blood pressure estimation by enhancing the PINN with the newly designed temporal blocks, which model precise personal temporal characteristics and address the scarcity of realistic healthy data.  
	 
	\item We propose a novel adversarial training method to augment physiological time series data for cuffless blood pressure estimation, which leverages physiological priors during the generation.
	
	\item We design a contrastive learning approach incorporated with adversarial training to capture blood pressure discriminative variations.
	
	\item Extensive experiments are carried out on graphene-HGCPT, ring-CPT, and blumio datasets, and the results show our methods' superiority in different modal physiological time series modeling using minimal ground truth data. 
\end{itemize}

\section{Related Work}
This section discusses the related work, including blood pressure monitoring, adversarial training, and contrastive learning methods.

\subsection{Blood Pressure Monitoring}
Blood Pressure (BP) is a key cardiovascular parameter commonly used by clinicians to evaluate cardiac and circulatory health~\cite{magwear}. Traditionally, BP is measured using cuff-based (auscultatory or oscillometric) digital BP devices, which compress the user's arm and often result in an unpleasant user experience~\cite{cuff1}. With the increasing use of contactless sensors for health monitoring~\cite{health-radio}, recent work tends to explore different modal cuffless BP measurement methods (\emph{e.g.}, bioimpedance, millimeter wave, PPG), which use wearable devices to capture pulse activities and infer BP using learning-based methods, like BiLSTM-based HNN~\cite{cao2021crisp}, and Transformer\cite{liang2023airbp, gonzalez2023benchmark}. However, they require amounts of invasive or cuff-based BP measurement results as ground truth labels to help map pulse activities to BP values. Particularly, beat-to-beat BP estimation and learning algorithms for continuous BP wave function extraction can be viewed as a time series modeling task, which is of significant importance in data mining and has attracted increasing attention in multiple areas~\cite{qi2019sports, qi2019stagnet, qi2020stc, qi2021semantics, qin2021deep, 9525836, qian2023towards, lv2024disentangled}.
Numerous models have been developed to analyze these beat-to-beat time series signals~\cite{ali2023efficient, barvik2021noninvasive}, which includes time domain LSTM-based methods~\cite{ali2023efficient}, a mixture of time and frequency domain human designed features-based approaches~\cite{barvik2021noninvasive}. These methods work well in data-sufficient situations, however, in precision physiological personal modeling, these models often fail due to limited training data and lack of personalized modeling. 
Unlike previous work, we propose to use a more generalized method for personal modeling which extracts intraperiodic features from the original waveform, and our proposed framework is based on physics constrain and personal modeling which only uses minimal training data while retaining high accuracy.   
\subsection{Adversarial Training}
Numerous studies have provided supporting evidence for the proposition that training with adversarial examples can effectively improve the capabilities of models~\cite{advprop, ilyas2019adversarial}. For instance, compared to clean examples, adversarial ones contribute to a more effective alignment of network representations with salient data characteristics and human perception~\cite{tsipras2018there}. 
Furthermore, models trained with adversarial examples demonstrate heightened robustness to high-frequency noise~\cite{yin2019fourier}. Besides, previous work~\cite{ivgi2021achieving} also demonstrates that by automatically generating adversarial examples, adversarial training will not decrease model performance but benefit both generalization and robustness of models. And He \emph{et al.}~\cite{he2023learning} use adversarial training to generate multivariate time-series samples for forecasting tasks. Especially for Physics-Informed Neural Networks~(PINNs), of which the robustness can be enhanced by fine-tuning the model with adversarial samples~\cite{li2023adversarial}. To the best of our knowledge, we are the first to adopt such an adversarial training method as data augmentation in BP estimation.

\subsection{Contrastive Learning}
Contrastive learning has shown remarkable success in deep learning tasks~\cite{10445106, 10113691}. Existing techniques like SimCLR~\cite{chen2020simple} and SimCSE~\cite{gao2021simcse} have demonstrated superior performance in generating semantically rich image embeddings without the need for labeled data. These methods help pull instances together with the same labels in the latent space, while simultaneously pushing clusters of samples from different classes apart. However, contrastive learning is rarely applied to time series analysis~\cite{spathis2022breaking}. Several common image augmentations including color changes or rotation may not be as relevant to time series data, especially in medical time series data mining. 
To address this insufficiency, we introduce contrastive learning to enhance the deep models' performance by capturing discriminative variations in cardiovascular dynamics. 

\section{Preliminaries}\label{sec: preliminary}

In this section, we start with the problem definition and then introduce the physics-informed neural networks for beat-to-beat BP estimation.

\subsection{Problem Definition}
Given the input time series signal $\mathbf{X}\in \mathbb{R}^{N\times T\times C}$, the goal of cuffless blood pressure is to estimate blood pressure value $y$ (\emph{e.g.}, systolic BP (SBP) or diastolic BP (DBP)). Let $\mathbf{X}=[\mathbf{x}_1, \mathbf{x}_2, \ldots, \mathbf{x}_N]$ 
be the continuous bioimpedance (BioZ) signal with $N$ segments, length-$T$ and $C$ channels, for each segment $\mathbf{x}_i \in \mathbb{R}^{T\times C}$, we extract physiological feature $\mathbf{u}_i \in \mathbb{R}^{M}$ denoted as the $M$ dimensional vector. The neural network can be formulated as $\hat{y} = f(\mathbf{x}, \mathbf{u}, \theta)$ with input $\mathbf{x}$, $\mathbf{u}$, parameter $\theta$, and the predicted blood pressure $\hat{y}$. 

\begin{figure*}[!ht]
	\centering
	\includegraphics[width=\textwidth]{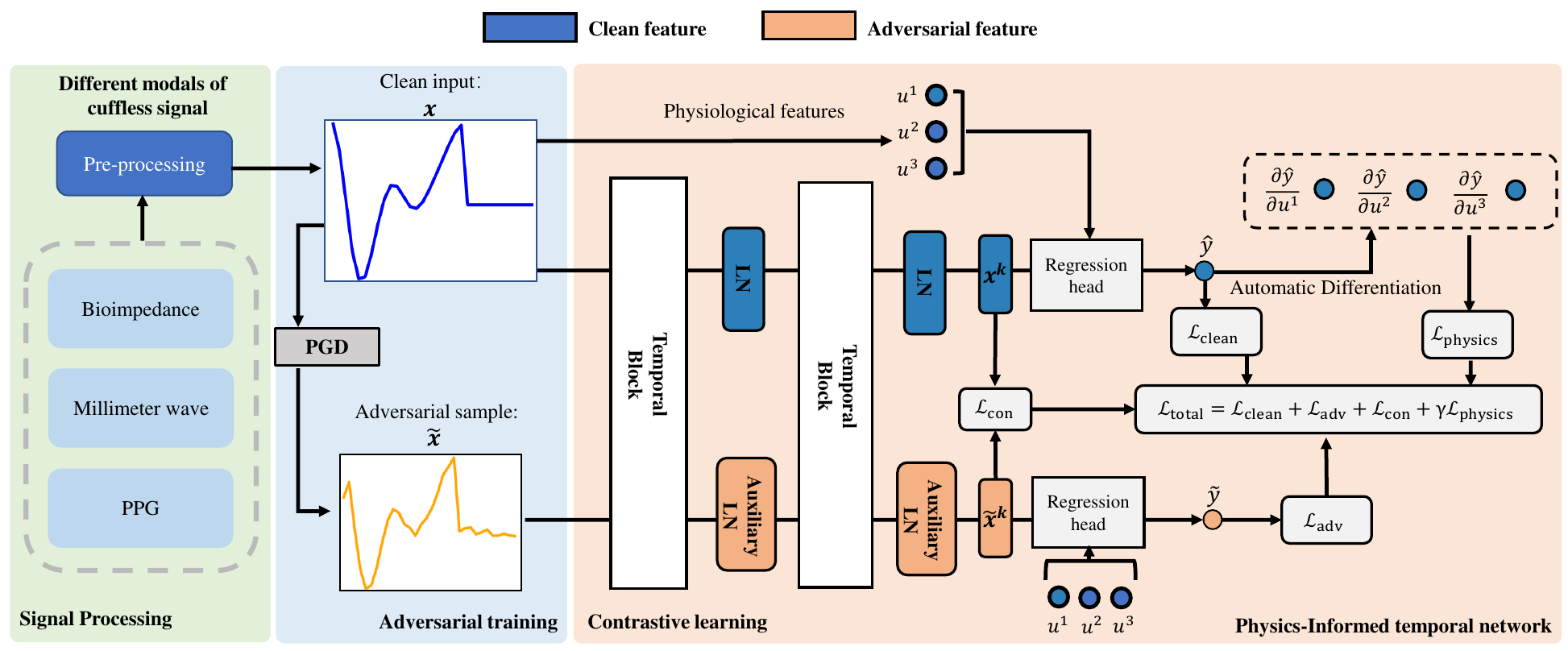} 
	\caption{Overall framework of physics-informed temporal network with adversarial contrastive learning for different modal cuffless blood pressure estimation. Our framework mainly contains two data flows: the upper part is about clean signals, which generate predicted values and differentiation for Taylor's approximation. The bottom part is adversarial samples generated by PGD accordingly. Notably, we introduce the adversarial training method in the PGD and construct the contrastive module to enable the model's ability in limited psychological data training. In the inference phase, the auxiliary LNs are not activated for different data distribution between clean and adversarial samples.}
	\label{fig: model overview}

\end{figure*}

\subsection{Physics-Informed Neural Networks} In this part, we review the formal definition of Physics-Informed Neural Networks (PINNs)~\cite{raissi2019physics}. PINNs are trained to solve regular supervised tasks while adopting several given laws of physics described by partial differential equations (PDEs) as the network residual. However, the relationships that connect wearable measurements to cardiovascular parameters are not well-defined in the form of generalized PDEs. Following~\cite{sel2023physics}, we adopt the idea of building physics constraints with Taylor’s approximation for certain gradually changing cardiovascular phenomena (\emph{e.g.}, establishing the relationship between physiological features extracted from bioimpedance sensor measurements and BP). 
Here we can define a polynomial with Taylor’s approximation around $i$-th segment as the following:
\begin{align}
	\Tilde{f}_i(\mathbf{x}, \mathbf{u}, \theta) = f(\mathbf{x}_i, \mathbf{u}_i, \theta) + \nabla_{\mathbf{u}_i}f(\mathbf{x}_i, \mathbf{u}_i, \theta)^T(\mathbf{u}-\mathbf{u}_i),
	\label{eq:taylor}
\end{align}%

\noindent where $\Tilde{f}_i(\mathbf{x}, \mathbf{u}, \theta)$ represents the above Talyor polynomial approximated based on $i$-th segment. The first-order Jacobin matrix can be calculated discretely as $\sum_{k=1}^M\frac{\partial f}{\partial \mathbf{u}^k}(\mathbf{u}^k-\mathbf{u}^k_i)$. 
A residual result from the difference between the neural network prediction and the Taylor polynomial evaluated at the $(i+1)$-th segment as shown in Eq.~(\ref{eq:taylor app}):
\begin{align}
	h_i(\mathbf{x}_{i+1}, \mathbf{u}_{i+1}, \theta) = \Tilde{f}_i(\mathbf{x}_{i+1}\mathbf{u}_{i+1}, \theta) -f(\mathbf{x}_{i+1}, \mathbf{u}_{i+1}, \theta),
	\label{eq:taylor app}
\end{align}
where $h_i(\mathbf{x}_{i+1}, \mathbf{u}_{i+1}, \theta)$ denotes the residual value evaluated at ($i+1$)-th segment using Taylor's approximation around $i$-th segment. The value of $h$ represents a physics-based loss for the neural network. Given that $h$ is calculated in an unsupervised way~(\emph{i.e.}, labels of output are not used), we can calculate $h$ for any given input sequence. We evaluate the value of $h$ for all consecutive input segments and use the mean squared sum of this evaluation for the physics-based loss function, as shown in Eq.~(\ref{eq:physics}):
\begin{align}
	\mathcal{L}_{\mathrm{physics}} = \frac{1}{(N-1)}\sum_{i=1}^{N-1}(h_i(\mathbf{x}_{i+1}, \mathbf{u}_{i+1}, \theta))^2, 
	\label{eq:physics}
\end{align}%
where $N$ is the total number of segments.

Following~\cite{sel2023physics}, as for physiological feature $\mathbf{u}$, we denote the amplitude change as $u^1$, which serves as a proxy for the extent of arterial expansion. The second feature is represented by $u^2$, which captures the inverse of the relative time difference between the forward-traveling (\emph{i.e.}, systolic) wave and the reflection wave, providing an estimate of the pulse wave velocity (PWV). The third feature denoted as $u^3$, which corresponds to beat-to-beat heart rate. These three physiological features can be formulated as follows:
\begin{align}
	u^1 = \Delta Z_{\mathrm{max}} - \Delta Z_{\mathrm{min}}, u^2 = \frac{1}{(t_F-t_B)}, u^3 = \frac{60}{t_J-t_A},
	\label{eq:physological features}
\end{align}%
where $\Delta Z_{\mathrm{max}}, \Delta Z_{\mathrm{min}}$ denotes the maximum and minimum variation in the (inverted) bioimpedance signal, $(t_F, t_B)$ refer to the trough point in the derivative signal and the zero-crossing point in the derivative signal transitioning to the ascent, and $(t_J, t_A)$ denote the peak point in the derivative signal, the zero-crossing point in the derivative signal transitioning to the descent, respectively.

\section{Proposed Approach}
We give an overview of our proposed framework in Section~\ref{sec: overview}. We explain each component of our framework including the PITN model with temporal blocks in Section~\ref{sec: PITN}, the adversarial examples generation in Section~\ref{sec: adversarial training}, the contrastive learning in Section~\ref{sec: contrastive learning}. Lastly, we discuss the training and inference process in Section~\ref{sec: training and inference}.

\subsection{Overview}\label{sec: overview}

As illustrated in Fig.~\ref{fig: model overview}, our framework comprises three key components: 1) Physics-Informed Temporal Network (PITN), 2) adversarial training, and 3) contrastive learning. In the PITN, newly designed temporal blocks are employed to extract personal physiological features, and Projected Gradient Descent (PGD) is tailed to generate adversarial samples for model training augmentation. These adversarial samples, along with clean samples, are disentangled using distinct layer normalization (LN) layers. Additionally, we adopt the contrastive learning loss by comparing clean and adversarial samples, based on their true blood pressure (BP) values. 

The whole process is as follows. We first pre-process different modal signals (\emph{e.g.} bioimpedance signal) to obtain clean input $\mathbf{x}$. Then adversarial sample $\Tilde{\mathbf{x}}$ is generated using Project Gradient Descent (PGD). Both clean and adversarial inputs are fed through the temporal block for feature extraction and the output estimated BPs are denoted as $\hat{y}$ and $\Tilde{y}$, respectively.

\subsection{Physics-Informed Temporal Networks}~\label{sec: PITN}

Physics-Informed Neural Networks (PINNs) traditionally use coordinates as input while overlooking potential temporal relationships among these inputs. This omission disregards the crucial inherent temporal dependencies present in practical physical systems, leading to a failure in globally propagating initial condition constraints and capturing accurate solutions across diverse scenarios~\cite{zhao2023pinnsformer}. In this paper, we introduce a novel temporal modeling block to enhance the capability of PINNs in modeling time-series medical data, by extracting complex temporal variations from transformed 2D tensors using a multi-scale Inception block.

\begin{figure}[ht]
	\centering
	\includegraphics[width=0.44\textwidth]{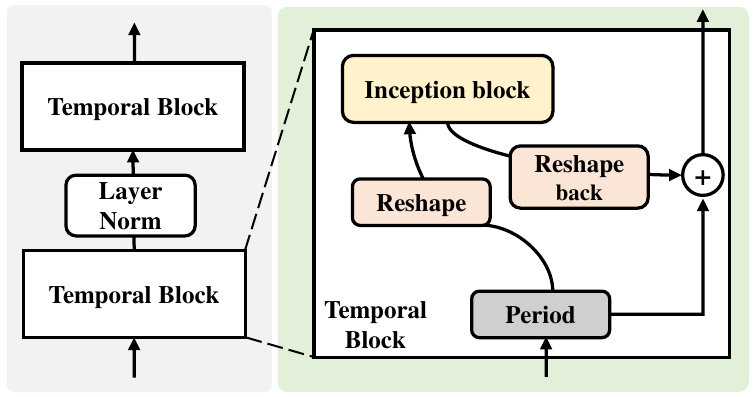} 
	\caption{Illustration of the proposed temporal block, which is designed to capture personalized cardiovascular cycles from the 1D signal by transforming it into a 2D tensor, followed by feature extraction using an inception block. The temporal block is stacked in a residual manner.} 
	\label{fig: temporal block}

\end{figure}

As illustrated in Fig.~\ref{fig: temporal block}, we organize the temporal block in a residual way~\cite{he2016deep}. Specifically, for the length-$T$ 1D input signal $\mathbf{x}\in \mathbb{R}^{T\times C}$, we project the raw inputs into the deep features $\mathbf{x}^0 \in \mathbb{R}^{d_{\mathrm{model}}}$ through the initial embedding layer, represented as $\mathbf{x}^0=\mathrm{Embed}(\mathbf{x})$. For the $k$-th layer of temporal block, the input $\mathbf{x}^{k-1} \in \mathbb{R}^{d_{\mathrm{model}}}$ is processed as:

\begin{equation}
	\mathbf{x}^k=\mathrm{TemporalBlock}(\mathbf{x}^{k-1})+\mathbf{x}^{k-1}. 	
	\label{eq: temporalblock}
\end{equation}

Each temporal block is designed to capture both intraperiod personal characteristics of physiological signals, as shown in Fig.~\ref{fig: temporal block}. This process includes operations such as  $\mathrm{Period}(\cdot)$, two $\mathrm{Reshape}(\cdot)$ functions, a 2D inception convolution $\mathrm{Inception}(\cdot)$ and residual addition. Initially, the 1D signal is extended to a 2D shape by calculating: 
\begin{equation}
	A = \mathrm{Avg}(\mathrm{Amp}(\mathrm{FFT}(\mathbf{x}))), f=\mathrm{argmax}(A), p=\lceil \frac{T}{f} \rceil,
	\label{eq: period}
\end{equation}

\noindent where $\mathrm{FFT}(\cdot)$ and $\mathrm{Amp}(\cdot)$ denote the Fast Fourier Transform and the calculation of amplitude values, respectively. $A$ represents the calculated amplitude for each frequency, averaged by $\mathrm{Avg}(\cdot)$. The personal cardiovascular cycle is identified by detecting the periodic basis function as described in Eq.~(\ref{eq: period}). For computational efficiency and avoidance of noises~\cite{zhou2022fedformer}, only one most significant frequency is selected for personalized modeling. The entire process described in Eq.~(\ref{eq: period}) can be summarized as $(A, f, p) = \mathrm{Period}(\mathbf{x})$. 	

For processing extended 2D signal, the temporal block is formulated as follows:
\begin{equation}
	\mathbf{X}_{\mathrm{2D}}^k = \mathrm{Reshape}_{p, f}(\mathrm{Padding}(\mathbf{x}^k)),
\end{equation}

\noindent where $\mathrm{Padding}(\cdot)$ extends the time series by zeros along the temporal dimension to ensure compatibility with $\mathrm{Reshape(\cdot)}$. For 2D signal encoding, $\mathrm{Inception(\cdot)}$ is employed, based on the InceptionNet~\cite{szegedy2015going}, which utilizes multi-scale 2D kernels, formulated as: 
\begin{equation}
	\hat{\mathbf{X}}_{\mathrm{2D}}^k=\mathrm{Inception}(\mathbf{X}_{\mathrm{2D}}^k),
\end{equation}

\noindent where $\mathbf{X}_{\mathrm{2D}}^k\in \mathbb{R}^{p\times f\times d_{\mathrm{model}}}$ is the transformed 2D tensor. The learned 2D representation $\hat{\mathbf{X}}_{\mathrm{2D}}^k$ is then transformed back to 1D space $\mathbf{x}^k \in \mathbb{R}^{d_{\mathrm{model}}}$ for subsequent stacked temporal blocks. After obtaining the temporal embeddings $\mathbf{x}^k$, the physiological features $\mathbf{u}$ are concatenated for regression. The estimated blood pressure value is then computed as follows:
\begin{equation}
    \hat{y} = \mathrm{RegHead}([\mathbf{x}^k \, || \, \mathbf{u}]),
\end{equation}
where $\mathrm{RegHead}(\cdot)$ represents a fully connected layer, and $||$ denotes concatenation. The embedding
$\mathbf{x}^k$ encapsulates rich temporal features from the sensor signal, which complements the PINN model. By integrating these two sets of features, precise blood pressure measurements can be achieved.

\subsection{Adversarial Training}~\label{sec: adversarial training}

Due to the limited availability of labeled physiological time series data, the results of current methods are still unsatisfactory. To address this issue, we enable blood pressure estimation using adversarial training by generating additional samples, while preserving the performance on clean data~\cite{advprop}.

For estimating BP with the function $f(\mathbf{x}, \mathbf{u}, \theta)$, we formulate the adversarial training as an optimization problem:
\begin{equation}\label{eq:optimization_problem}
	\begin{aligned}
		\max_{\Delta}& \quad \mathcal{L}(f(\mathbf{x}+\Delta, \mathbf{u}, \theta), f(\mathbf{x}, \mathbf{u}, \theta)) \\
		\mathrm{s.t.} & \quad ||\Delta||_{\infty} \leq \epsilon,
	\end{aligned}
\end{equation}

\noindent where $\mathcal{L}(\cdot)$ represents the mean square error between $f(\mathbf{x}+\Delta, \mathbf{u}, \theta)$ and $f(\mathbf{x}, \mathbf{u}, \theta)$, and $\epsilon$ is the threshold for the maximum allowable adversarial perturbation. The perturbation is initialized drawn from a uniform distribution $\Delta \sim \mathcal{N}(0, \sigma^2)$.
Subsequently, the adversarial perturbation is updated as $ \Delta = \eta \times \mathrm{sign}\nabla_\mathbf{x}f(\mathbf{x}, \mathbf{u}, \theta)$, where $\mathrm{sign}$ is the element-wise sign function, producing $-1$ for negative values of the Jacobin matrix and $+1$ otherwise. $\eta$ denotes the step size of each iteration.

In our proposed blood pressure estimation framework, we further regularize the perturbed signal by truncating it within the problem domain $\Omega$: 
\begin{equation}\label{eq:update adversarial perturbation}
	\begin{aligned}
		\mathbf{\Tilde{x}} = \mathrm{clip}_{\Omega}(\mathbf{x} + \Delta),
	\end{aligned}
\end{equation}

\noindent 
where the problem domain is defined by the element-wise maximum and minimum values of the input, $\Omega = [\mathrm{min}(\mathbf{x}), \mathrm{max}(\mathbf{x})]$ for all training data. We adopt the Projected Gradient Descent (PGD)~\cite{madry2018towards} in time series adversarial samples generating, as shown in Algorithm~\ref{algorithm:1}. During initialization, the clean BioZ data $\mathbf{x}$ is perturbed with randomly sampled variables $\Delta$ from the uniform distribution. The $\mathrm{clip}_{\Omega}(\cdot)$ means clipping samples according to the problem domain $\Omega$. During the projected gradient descent process, gradient ascent is applied to maximize the residual, and the perturbed sample is subsequently truncated again to remain within the problem domain.

\begin{algorithm}[h]
	\caption{Physiological time series data generating by PGD}
	\textbf{Input}: Clean BioZ data $\mathbf{x}$, problem domain $\Omega$, proposed model $f(\mathbf{x}, \mathbf{u}, \theta)$. \\
	\textbf{Parameter}: Adversarial perturbation $\Delta \sim \mathcal{N}(0, \sigma^2)$, number of iteration steps $I$, and the step size of each iteration $\eta$. \\
	\textbf{Output}: Adversarial samples $\Tilde{x}$. \\
	\vspace{-4mm}
	\begin{algorithmic}[1] 
		\FOR{$t=1:I$} \do
		\STATE 
		\STATE $\mathbf{\Tilde{x}} = \mathbf{x} + \Delta$ \hfill // Perturbation
		\STATE $\mathbf{\Tilde{x}} = \mathrm{clip}_{\Omega}(\mathbf{\Tilde{x}})$ 
		\STATE $ \Delta = \eta \times \mathrm{sign}(\nabla_\mathbf{x}f(\mathbf{x}, \mathbf{u}, \theta))$ \hfill // Gradient ascent
		\STATE $\mathbf{\Tilde{x}} = \mathbf{\Tilde{x}} + \Delta$
		\STATE $\mathbf{\Tilde{x}} = \mathrm{clip}_{\Omega}(\mathbf{\Tilde{x}})$
		\ENDFOR
		\STATE \textbf{return} $\mathbf{\Tilde{x}}$
	\end{algorithmic}
	\label{algorithm:1}
\end{algorithm}

\subsection{Contrastive Learning}\label{sec: contrastive learning}

Although generating additional samples is beneficial for augmenting cuffless blood pressure data, increasing data points can lead to fluctuations in per-subject correlation~\cite{sel2023physics}. Recently, contrastive learning has achieved success in self-supervised learning~\cite{demirel_finding_nodate}, by introducing additional constraints in the latent space to regularize embedding with the same label. Specifically, in classification tasks, similar data pairs (referred to as positive pairs) are selected for the same class, while all other pairs are treated as negative ones. However, applying contrastive learning to regression tasks remains non-trivial. In this part, we first introduce contrastive learning into the BP regression issue. Here, we define the $i$-th and $j$-th samples as a positive pair based on the following criterion:
\begin{align}
	y_{\mathrm{shift}} \geq |y_i-y_j|,
	\label{eq:y_shift}
\end{align}

\noindent where $y_{\mathrm{shift}}$ is a threshold parameter. By imposing this soft constraint, which labels inputs with similar blood pressure (BP) values, we aim to enhance the model's ability to learn cardiovascular dynamics from a single pulse wave. The contrastive learning loss is computed between clean and adversarial samples. Inspired by~\cite{khosla2020supervised}, we formalize the contrastive learning loss as follows:
\begin{align}
	\mathcal{L}_{\mathrm{con}} = \sum_{i\in S}\frac{-1}{|P(i)|}\sum_{p\in P(i)} log\frac{\mathrm{exp}(\mathbf{x}_i^k \cdot \mathbf{x}_p^k / \tau)}{\sum_{a\in A(i)} \mathrm{exp}(\mathbf{x}_i^k \cdot \mathbf{x}_a^k / \tau)},
	\label{eq:loss contrastive}
\end{align}

\noindent where $i \in S$ is the index of an arbitrary training sample, and index $p$ denotes an adversarial sample which has $|y_p-y_i| < y_{\mathrm{shifit}}$. Index $p$ is called positive and index $a$ is called negative.
$P(i)$ and $A(i)$ are sets of positive samples and negative samples, respectively.
The embeddings $\mathbf{x}_p^k$ and $\mathbf{x}_a^k$ correspond to positive/negative pairs, respectively, after $k$ layers of Temporal blocks for the input $\mathbf{x}_i$. The symbol $\cdot$ represents the inner (dot) product, and $\tau$ is a scalar temperature parameter.

\subsection{Training and Inference}\label{sec: training and inference}

\subsubsection{Training}

We formally integrate the Projected Gradient Descent (PGD) method into the regression framework, as outlined in Algorithm~\ref{algorithm:1}. For each clean input, we initiate the process by using PGD to generate its corresponding adversarial counterpart. Subsequently, both the clean data and its adversarial counterpart are fed into the same network but with different normalization layers~(LNs) inspired by~\cite{advprop}. Specifically, the primary LNs are applied to the clean input, while the auxiliary LNs are used for the adversarial counterpart. The overall loss is then minimized with respect to the network parameters through gradient updates. It is important to note that, aside from the LN layers, all other layers are optimized concurrently for both clean inputs and their adversarial counterparts.

During the training phase, both clean and adversarial samples are input into the network to obtain blood pressure predictions. The regression loss $\mathcal{L}_{\mathrm{clean}}$ and $\mathcal{L}_{\mathrm{adv}}$, are calculated as follows:
\begin{align}
	\begin{aligned}
		\mathcal{L}_{\mathrm{clean}} &= \frac{1}{S}\sum^S_{i=1}(\hat{y}_{i} - y_i)^2, \\
		\mathcal{L}_{\mathrm{adv}} &= \frac{1}{S}\sum^S_{i=1}(\Tilde{y}_{i} - y_i)^2,
	\end{aligned}
	\label{eq:combined loss}
\end{align}

\noindent where $S$ denotes the total number of training samples, and $\hat{y}_i$ and $\Tilde{y}_i$ correspond to the network outputs for the $i$-th clean example $\mathbf{x}_i$ and adversarial example $\Tilde{\mathbf{x}}_i$, respectively. The overall loss is computed by summing the regression loss from Eq.~(\ref{eq:combined loss}), the physical loss from Eq.~(\ref{eq:physics}), and the contrastive loss from Eq.~(\ref{eq:loss contrastive}). The total loss can be formulated as the following:
\begin{align}
	\mathcal{L}_{\mathrm{total}} = \mathcal{L}_{\mathrm{clean}} + \mathcal{L}_{\mathrm{adv}} + \mathcal{L}_{\mathrm{con}} + \gamma \mathcal{L}_{\mathrm{physics}},
\end{align}%

\noindent where $\gamma$ is a hyper-parameter which is set to 1 for balancing the magnitude of multiple losses.

\subsubsection{Inference} Given a test bioimpedance segment $\mathbf{x} \in \mathbb{R}^{T \times C}$ representing an entire cardiac cycle, along with physiological features $\mathbf{u} \in \mathbb{R}^{M}$, the network output $\hat{y} = f(\mathbf{x}, \mathbf{u}, \theta)$ is obtained using primary layer normalizations (LNs). The output corresponds to the blood pressure (BP) value (e.g., systolic BP or diastolic BP).

\section{Experiments}

In this section, we comprehensively evaluate the performance of our proposed framework on three public benchmark datasets, containing various wearable signals for cuffless blood pressure estimation. The evaluation aims at estimating systolic BP (SBP) and diastolic BP (DBP). The datasets are introduced first, followed by a description of the baselines used for comparison, and a brief outline of the evaluation protocols and implementation details. Then, the experimental results are presented with the corresponding analysis. Finally, we conduct ablation studies on each module proposed in this paper.

\subsection{Datasets}

We conduct our experiments on Graphene-HGCPT~\cite{kireev2022continuous}, Ring-CPT~\cite{sel2023continuous} and Blumio~\cite{blumio} datasets. The first two datasets contain raw time series data obtained from a wearable bioimpedance sensor, with corresponding reference BP values acquired using a medical-grade finger cuff (Finapres NOVA). The Blumio dataset includes data from 115 subjects (aged range 20-67 years), collected using several types of wearable sensors: PPG, applanation tonometry, and the Blumio millimeter-wave radar. Detailed information on each dataset is provided below.

\subsubsection{Graphene-HGCPT Dataset~\cite{kireev2022continuous}}
This BioZ signal based dataset includes data from six participants who undergo multiple sessions of a blood pressure elevation routine, which involves hand grip (HG) exercise followed by a cold pressor test (CPT) and recovery. Participants wear bioimpedance sensors that utilize graphene e-tattoos placed on their wrists along the radial artery. They also wear a silver electrode-based wristband at different positions. The evaluation takes over 24,829 samples (after post-processing), covering a wide range of BP values.

\subsubsection{Ring-CPT Dataset~\cite{sel2023continuous}}
This dataset consists of data from five participants who undergo multiple sessions of CPT and recovery. Bioimpedance data is collected with a ring-worn BioZ sensor placed on the participants’ fingers.
Our experiments adhere to a minimal training criterion~\cite{sel2023physics}, where the labeled data is divided into bins, with $S = \mathrm{BP}_{\mathrm{range}}\times 2$. Here, $\mathrm{BP}_{\mathrm{range}}$ represents the difference between the maximum and minimum BP values, calculated separately for SBP and DBP. This binning process partitions the dataset into distinct bins with a bin width of 0.5 mmHg, from which one data point is randomly selected from each bin to form the initial training set. The overall evaluation takes place over the total 6,544 samples for Ring-CPT. 

\subsubsection{Blumio Dataset~\cite{blumio}}
This multimodal dataset includes data from 115 subjects using multiple wearable sensors. In our experiments, we select a subset of 30 subjects, aged 21 to 50 years (15 males and 15 females), and test our framework on PPG and millimeter wave signals to evaluate our methods across different modalities for cuffless BP estimation. For data processing, we employ the same procedure as for the Ring-CPT and Graphene-HGCPT datasets.

We apply a consistent preprocessing method across all three datasets, extracting beat-to-beat signals from a single channel to ensure that our approach remains universal across different datasets and modalities. This approach differs slightly from the method used in~\cite{sel2023physics}, which integrates multiple channels in the graphene-HGCPT dataset. For each of the three datasets, separate models are trained for each subject to achieve precise, personalized modeling. Each model receives a varying number of labeled training points according to the minimal training criterion. Subsequently, we evaluate the performance of each model against reference blood pressure (BP) values using a test set, which includes BP values not used in the training process.

%

\subsection{Baselines}

We compare our model against several baseline models and state-of-the-art approaches including methods specifically designed for cuffless blood pressure estimation as Hybrid-LSTM~\cite{schrumpf2021assessment}, Physics-Informed Neural Networks (PINN)~\cite{sel2023physics} and ResNet1D~\cite{gonzalez2023benchmark}. Additionally, we evaluate our framework against several general time series models, such as inverted Transformer (iTransformer)~\cite{liu_itransformer_2024} and TimesNet~\cite{wu_timesnet_2023}. We follow the methodologies provided by each corresponding paper during the experiment. More details refer to the supplementary.

\subsection{Evaluation Protocols} 

Following the standard practice in blood pressure estimation~\cite{liang2023airbp, sel2023physics}, we adopt common evaluation metrics, including root-mean-square error (RMSE), and Pearson’s correlation coefficient values. RMSE is defined as follows:
\begin{equation}
	\mathrm{RMSE} = \sqrt{\frac{1}{N-S}\sum_{i=1}^{N-S}(\hat{y}_i-y_i)^2},
\end{equation}
where $\hat{y}_i$ and $y_i$ correspond to $i$-th estimated and true BP value (SBP or DBP), and $N-S$ is the total number of test samples.
Pearson's correlation coefficient $r$ is calculated to measure the linear correlation between the estimated and true BP values, defined as:
\begin{equation}
	r = \frac{\sum_{i=1}^{N-S}(\hat{y}_i - \bar{Y})(y_i - \bar{Y}_{true})}{\sqrt{\sum_{i=1}^{N-S}(\hat{y}_i - \bar{Y})^2 \sum_{i=1}^{N-S}(y_i - \bar{Y}_{true})^2}},
\end{equation}
where $\bar{Y}$ and $\bar{Y}_{\mathrm{true}}$ denote the mean values of the predicted ($\hat{y}$) and true BP values ($y$), respectively.

Additionally, the mean error (ME) and standard deviation of the error (SDE) are computed following the American National Standards Institute/ Association for the Advancement of Medical Instrumentation/ International Organization for Standardization AAMI standard. This standard requires BP devices to have ME and SDE values less than 5 and 8 mmHg, respectively~\cite{stergiou2018universal}. Moreover, we utilize a pair-wise t-test~\cite{ruxton2006unequal} to compare the performance of our proposed method and other baseline models, in order to make the experimental results more convincing. More results under the AAMI standard and statistical significance analyses refer to the supplementary.

\subsection{Implementation Details}

 All implementations are based on the open-source PyTorch framework and trained on a single NVIDIA 3090 GPU. The networks are trained using the Adam optimizer with a learning rate set to 0.001. For generating controllable adversarial samples, we choose a relatively small perturbation parameter $\epsilon$ of 0.2 and set the training steps to 2. During training, we balance the magnitude of different losses by setting hyperparameters based on a parameter sensitivity analysis (see supplementary material). Specifically, we set the weight of physics loss to $\gamma=1$ and the minimal sensitivity of blood pressure to $y_{\mathrm{shift}}=2$.

\subsection{Results and Analysis}\label{sec: results and analysis}

\begin{table*}[ht]
	\centering
		\caption{Results comparison for BP estimation on Graphene-HGCPT dataset in terms of our proposed model and other state-of-the-art methods. \emph{Ours-Full} and \emph{Ours-Base} refer to our Full model and our model without adversarial training and contrastive learning, respectively.
        $\uparrow$ refers to the higher result being better and $\downarrow$ vice versa. The bold values represent the best performance and underlined values indicate the second best performance.}
	\resizebox{\textwidth}{!}{%
		\begin{tabular}{cc cc cc cc  cc cc cc cc}
			\toprule
			\multicolumn{2}{c}{\multirow{2}{*}{Models}} &
			\multicolumn{2}{c}{iTransformer} &
			\multicolumn{2}{c}{TimesNet} &
			\multicolumn{2}{c}{ResNet1D} &
			\multicolumn{2}{c}{Hybrid-LSTM} &
			\multicolumn{2}{c}{PINN} &
		 	\multicolumn{2}{c}{Ours-Base} &
			 \multicolumn{2}{c}{Ours-Full} 
			\\
			\multicolumn{2}{c}{} &
			\multicolumn{2}{c}{2024} &
			\multicolumn{2}{c}{2023} &
			\multicolumn{2}{c}{2023} &
			\multicolumn{2}{c}{2022} &
			\multicolumn{2}{c}{2023} &
			\multicolumn{2}{c}{2024} &
			\multicolumn{2}{c}{2024}

			\\
			\cmidrule{3-16}
			
			\multicolumn{2}{c}{BP type} & 
			Corr $\uparrow$ & RMSE $\downarrow$ & Corr $\uparrow$& RMSE $\downarrow$& Corr $\uparrow$ & RMSE $\downarrow$ & Corr $\uparrow$ & RMSE $\downarrow$ & Corr $\uparrow$ & RMSE $\downarrow$ &
			Corr $\uparrow$ & RMSE $\downarrow$ & Corr $\uparrow$ & RMSE $\downarrow$  \\
			\midrule
			\multirow{7}{*}{\rotatebox{90}{SBP}} 
			& 1   &0.41  &16.5  &0.38  & 11.5  & 0.18 & 13.3 &0.35  & 20.1 & \underline{0.66} & 9.5 & 0.63 & \underline{9.5} & \textbf{0.68} & \textbf{9.0}  \\
			& 2   &0.43  &19.1  &0.07  &12.3  & 0.36 & 17.7 &0.36  & 24.9 & 0.48 & 11.1  & \underline{0.61} & \underline{9.6} & \textbf{0.62} & \textbf{9.5}  \\
			& 3   &0.48  &17.0  &0.40  &12.9  & 0.16 & 15.5 & 0.42 & 16.8 & 0.43 & 12.6 & \underline{0.60} & \underline{10.7} & \textbf{0.61} & \textbf{10.6}  \\
			& 4   &0.42  &17.2  &0.45  &15.7   & 0.23 & 19.6 & 0.35 & 23.2 & 0.48 & 14.4& \underline{0.57}  & \underline{13.0} & \textbf{0.59}  & \textbf{12.8}\\
			& 5   &0.27  &18.6 &0.30  &13.6  &0.28  & 17.7 & 0.20 & 22.3  & 0.47 & 12.1 & \underline{0.60} & \underline{10.6}  & \textbf{0.62}  & \textbf{10.6} \\
			& 6   &0.24  &27.1  &0.30  &20.5  & 0.15 & 21.5 & 0.21 & 34.1 & 0.39 & 17.2  & \underline{0.59} & \underline{14.9} & \textbf{0.68} & \textbf{13.3} \\
			\cmidrule{2-16}
			& Avg &0.38  &19.3  &0.32  &14.4   & 0.23 & 17.6 & 0.32 & 23.6 & 0.48 & 12.8 &\underline{0.60} & \underline{11.4} & \textbf{0.63} & \textbf{11.0} \\
			\midrule
		
			\multirow{7}{*}{\rotatebox{90}{DBP}} 
			& 1   &0.43  &16.5  &0.23  &10.8    & 0.49 & 12.3 & 0.42 & 18.3 & \underline{0.62} & \textbf{8.3}  & \textbf{0.62}  & \underline{8.6}  & 0.61 & 8.7 \\
			& 2   &0.34  &20.5  & 0.24 &11.2   & 0.27 & 13.3 & 0.31 & 22.5 &0.44  &9.8  & \underline{0.54} & \underline{9.1} & \textbf{0.54} & \textbf{9.2} \\
			& 3   &0.44  &15.2  &0.38  &11.4   & 0.21 & 10.3 & 0.32 & 16.6 &0.51  & 9.8  & \underline{0.64} & \underline{8.9} & \textbf{0.66} & \textbf{8.6}   \\
			& 4   &0.41  &16.2  & 0.23 &13.6  & 0.32 & 15.5 & 0.30 & 21.8 & \underline{0.50}  &12.8   & 0.47  & \underline{11.9} & \textbf{0.53} & \textbf{11.6}  \\
			& 5   &0.26  &20.0  &0.33  &15.8  & 0.33 & 19.6 & 0.17 & 26.3 &0.60  &11.9 & \underline{0.71} & \underline{9.9} & \textbf{0.74} & \textbf{9.4}   \\
			& 6   &0.18  & 22.5 &0.14  &15.5   & 0.10 & 17.7 & 0.17 & 30.3 & 0.31  & 14.5 & \textbf{0.37} & \underline{13.5} & \underline{0.34} & \textbf{13.3} \\
			\cmidrule{2-16}
			& Avg &0.34  &18.4  &0.26  &13.1   & 0.29 & 21.5 & 0.28 & 22.6 &0.50  &11.1 & \underline{0.56}  & \underline{10.3} & \textbf{0.57} & \textbf{10.1}   \\
		  \midrule
                \multicolumn{2}{c}{Inference time (s)}& \multicolumn{2}{c}{0.03} & \multicolumn{2}{c}{0.17} & \multicolumn{2}{c}{0.04} & \multicolumn{2}{c}{0.08} & \multicolumn{2}{c}{0.02} &  \multicolumn{2}{c}{0.18} & \multicolumn{2}{c}{0.18}          \\		
	
			\bottomrule
		\end{tabular}%
	
	}
	\label{tab:model_comparison Graphene-HGCPT}

\end{table*}

\begin{table*}[ht]
	\centering
	\caption{Results comparison for BP estimation on Ring-CPT dataset in terms of our proposed model and other state-of-the-art methods. \emph{Ours-Full} and \emph{Ours-Base} refer to our Full model and our model without adversarial training and contrastive learning, respectively. $\uparrow$ refers to the higher result being better and $\downarrow$ vice versa. The bold values represent the best performance and underlined values indicate the second best performance.}
	\resizebox{\textwidth}{!}{%
		\begin{tabular}{cc cc cc cc cc cc cc cc}
			\toprule
			\multicolumn{2}{c}{\multirow{2}{*}{Models}} &
			\multicolumn{2}{c}{iTransformer} &
			\multicolumn{2}{c}{TimesNet} &
			\multicolumn{2}{c}{ResNet1D} &
			\multicolumn{2}{c}{Hybrid-LSTM} &
			\multicolumn{2}{c}{PINN} &
			\multicolumn{2}{c}{Ours-Base} &
			\multicolumn{2}{c}{Ours-Full} 
			
			\\
			\multicolumn{2}{c}{} &
			\multicolumn{2}{c}{2024} &
			\multicolumn{2}{c}{2023} &
			\multicolumn{2}{c}{2023} &
			\multicolumn{2}{c}{2022} &
			\multicolumn{2}{c}{2023} &
			\multicolumn{2}{c}{2024} &
			\multicolumn{2}{c}{2024} 
			\\
			\cmidrule{3-16}
			\multicolumn{2}{c}{BP type} & 
			Corr $\uparrow$ & RMSE $\downarrow$ & Corr $\uparrow$ & RMSE $\downarrow$ & Corr $\uparrow$ & RMSE $\downarrow$ & Corr $\uparrow$ & RMSE $\downarrow$ & Corr $\uparrow$ & RMSE $\downarrow$ & Corr $\uparrow$ & RMSE $\downarrow$ &Corr $\uparrow$ & RMSE $\downarrow$   \\
			\midrule
	
			\multirow{6}{*}{\rotatebox{90}{SBP}}      
			& 1   &0.62  &10.5  &0.67  &8.8   &0.58  & 10.7 & 0.50 & 14.2 & 0.58  & 9.0 & \underline{0.71}  & \underline{7.6} & \textbf{0.73}  & \textbf{7.4} \\
			& 2   &0.47  &10.1  &0.43  &7.4  & 0.39 & 9.0 & 0.40 & 12.0 & 0.51  & 7.1  & \underline{0.66} & \underline{6.2} & \textbf{0.72}& \textbf{5.8} \\
			& 3   &0.43  &14.9  &0.42  &11.3  & 0.60 & 11.2 & 0.48 & 17.1 & 0.65 & 9.8  & \underline{0.67} & \underline{9.0} & \textbf{0.69} & \textbf{8.8} \\
			& 4  &0.71  &12.3  &0.79  &9.8 & 0.80 & 12.2 & 0.69 & 14.8 & 0.87 & 10.0  & \underline{0.90}  & \underline{6.5} & \textbf{0.90} & \textbf{6.5} \\
			& 5  &0.54  &8.8  &0.50  &8.1   & 0.56 & 8.8 & 0.49 & 11.5 & 0.68 & 7.4  & \underline{0.78} & \underline{5.7}  & \textbf{0.82} & \textbf{5.2}\\
			\cmidrule{2-16}
			& Avg & 0.55 &11.3  &0.56  &9.1  &0.59 & 10.4 & 0.51 & 13.9 & 0.66  & 8.7 & \underline{0.74}  & \underline{7.0} & \textbf{0.77} & \textbf{6.7}\\
			
			\midrule
			\multirow{6}{*}{\rotatebox{90}{DBP}}       
			& 1   &0.45  &11.4 &0.47  & 7.7  & 0.43 & 10.5  &0.49  & 13.6 &0.57  & 6.2 &\underline{0.66} & \underline{6.0} & \textbf{0.68}  & \textbf{5.7} \\
			& 2   &0.36  &9.4  &0.20  & 6.8  &  0.38&5.8  &0.21  & 12.9 &0.46  & \underline{4.8} & \underline{0.46} & 5.5 & \textbf{0.53} & \textbf{4.8}   \\
			& 3   &0.55  &10.3  &0.65  &7.7  & 0.68 & 10.7 &  0.52  & 12.0 &0.74  & 7.5 & \underline{0.78}  & \underline{6.4} & \textbf{0.79}  &  \textbf{6.1}  \\
			& 4  &0.69  &10.4  &0.62  &8.7   &  0.76& 12.9 & 0.67 & 12.2 &0.81  & 6.4& \underline{0.84} & \underline{5.1} & \textbf{0.84}  & \textbf{5.1}  \\
			& 5  &0.38  &8.3  &0.53  &5.6   & 0.54 & 6.6 & 0.36 & 11.3 & \underline{0.72}  & 4.9 & 0.71 & \underline{4.6} & \textbf{0.76}  & \textbf{4.3}  \\
			\cmidrule{2-16}
			& Avg &0.49  &10.0  &0.49  &7.3   & 0.56 & 9.3 & 0.45 & 12.4 &0.67  & 6.0 & \underline{0.69} & \underline{5.5} & \textbf{0.72} & \textbf{5.2}\\
              
			\bottomrule
		\end{tabular}%
	}
	\label{tab:model_comparison Ring-CPT}
\end{table*}

\subsubsection{Results on Graphene-HGCPT Dataset} 

Table~\ref{tab:model_comparison Graphene-HGCPT} presents a comprehensive overview of the quantitative outcomes obtained through our proposed framework and alternative baseline approaches. From the table, it is evident that the Ours-Full model demonstrates superior performance against other methods (\emph{i.e.}, iTransformer, TimesNet, ResNet1D, Hybrid-LSTM) in both SBP and DBP estimation. For example, the Ours-Full model achieves a 31\%/14\% improvement in correlation and a 14\%/9\% reduction in RMSE (SBP/DBP) compared to the PINN model, mostly attributing to the superiority and effectiveness of our proposed temporal block for personal cardiovascular cycle modeling. Additionally, the improved results of the Ours-Full model compared to the Ours-Base model show that adversarial examples and contrastive learning are beneficial for improving blood pressure estimation.

Noting that under minimal training criteria, strong baselines (such as ResNet or Transformer-based methods) also struggle to make precise predictions and model BP dynamics, which is possibly due to their pure data-driven nature. Models with prior knowledge designed for time series can improve performance, thus TimesNet and PINN model achieves better results than iTransformer, ResNet1D, and Hybrid-LSTM. However, compared to our methods, TimesNet lacks physiological priors, while PINN models are insufficient for extracting temporal information. Our model combines physiological priors with advanced temporal modeling, enabling more precise blood pressure estimation.

Given the real-time requirements of continuous BP estimation, we conduct experiments to evaluate inference time, as shown in the last row of Table~\ref{tab:model_comparison Graphene-HGCPT}. The results indicate that PINN and iTransformer demonstrate relatively fast inference times per segment, whereas TimesNet and our proposed models exhibit a bit slower inference speeds. This slower performance is primarily due to the transformation operations and the extraction of multi-scale temporal information. However, considering the obtained superior performance, our proposed model can achieve a good trade-off. 

Furthermore, the results underscore the varying challenges in modeling different subjects, as illustrated by the performance of subjects \#4 and subject \#6 in Table~\ref{tab:model_comparison Graphene-HGCPT}, which show higher RMSE and lower correlation compared to other subjects. This highlights the critical importance of personalized modeling. Our model employs temporal blocks specifically designed for temporal modeling, which better capture personalized cardiovascular cycles and enhance the accuracy of personal modeling.

\subsubsection{Results on Ring-CPT Dataset}

The results and comparisons of our model and other state-of-the-art methods on the ring-CPT dataset can be found in Table~\ref{tab:model_comparison Ring-CPT}. From the table, we can see that the Ours-Full model manifests the best performance against other methods by a considerable margin. This improvement is largely attributed to our model's capability in temporal modeling and the effectiveness of our data augmentation methods, including adversarial examples and contrastive learning.

\subsubsection{Results on Blumio Dataset} 

We report the results on the Blumio dataset to demonstrate our model is versatile and applicable to various modalities of cuffless blood pressure signals, such as mmWave and PPG. Table~\ref{tab: different model processing} presents the performance of our proposed approaches for SBP and DBP estimation using different modalities in the Blumio dataset (mmWave and PPG). From the table, it is evident that our proposed framework consistently achieves superior performance compared to other methods in terms of RMSE and Pearson’s correlation. Specifically, when using mmWave for BP estimation, the Ours-Full model achieves a 132\%/62\% improvement in correlation (SBP/DBP) compared to the TimesNet. Similarly, for BP estimation using PPG, the Ours-Full model shows a 100\%/56\% improvement in correlation (SBP/DBP) over the TimesNet. The enhanced performance of the Ours-Full model on Pearson's correlation is mostly due to the physiological prior and contrastive learning. However, the correlation results achieved in the Blumio using either PPG or mmWave modalities are generally inferior to those in the graphene-HGCPT and Ring-CPT datasets, which may stem from the sparse sampling points, causing inconsistent predictions.



\begin{table}[h]
	\centering
	\caption{Results comparison of different modal (mmWave and PPG) cuffless signal blood pressure estimation on Blumio dataset. \emph{Ours-Full} and \emph{Ours-Base} refer to our Full model and our model without adversarial training and contrastive learning, respectively. $\uparrow$ refers to the higher result being better and $\downarrow$ vice versa. The bold values represent the best performance and underlined values indicate the second best performance.}
	\label{tab:different modal}
	\resizebox{0.47\textwidth}{!}{%
		\begin{tabular}{ccccc}
			\toprule
			\multirow{3}{*}{Methods} & \multicolumn{2}{c}{mmWave} & \multicolumn{2}{c}{PPG} \\
			\cmidrule{2-5}
			& \multicolumn{2}{c}{SBP/DBP} & \multicolumn{2}{c}{SBP/DBP}  \\
			\cmidrule{2-5}
			& Corr $\uparrow$      & RMSE $\downarrow$  & Corr $\uparrow$      & RMSE $\downarrow$       \\
			\cmidrule{1-5}      
			Hybrid-LSTM         &  0.27/0.26    & 7.0/6.1 & 0.24/0.27 & 7.9/7.3  \\
			ResNet1D         &  0.29/0.30    & 6.7/6.0  & 0.28/0.35 & 7.2/5.9 \\
			iTransformer         &  0.27/0.27    & 6.6/5.2 & 0.28/0.32 & 7.5/5.7  \\
			TimesNet & 0.22/0.28  &  4.1/3.6  & 0.22/0.27 & 4.3/3.7  \\
			PINN  &  0.45/0.37    &  3.8/3.5 & 0.41/0.40 & \underline{4.0}/\underline{3.7} \\
			\cmidrule{1-5}
			Ours-Base  &  \underline{0.47}/\underline{0.40} &  \underline{3.8}/\underline{3.4}  & \underline{0.42}/\underline{0.40}  & 4.2/3.8     \\
			Ours-Full  &  \textbf{0.51}/\textbf{0.44} &  \textbf{3.7}/\textbf{3.3}  & \textbf{0.44}/\textbf{0.42}  &  \textbf{3.5}/\textbf{3.7}    \\
			
			\midrule
		\end{tabular}%
	}
	\label{tab: different model processing}
\end{table}

\subsubsection{Qualitative Results Analysis} 
\begin{figure*}[ht]
	\centering
	\includegraphics[width=0.9\textwidth]{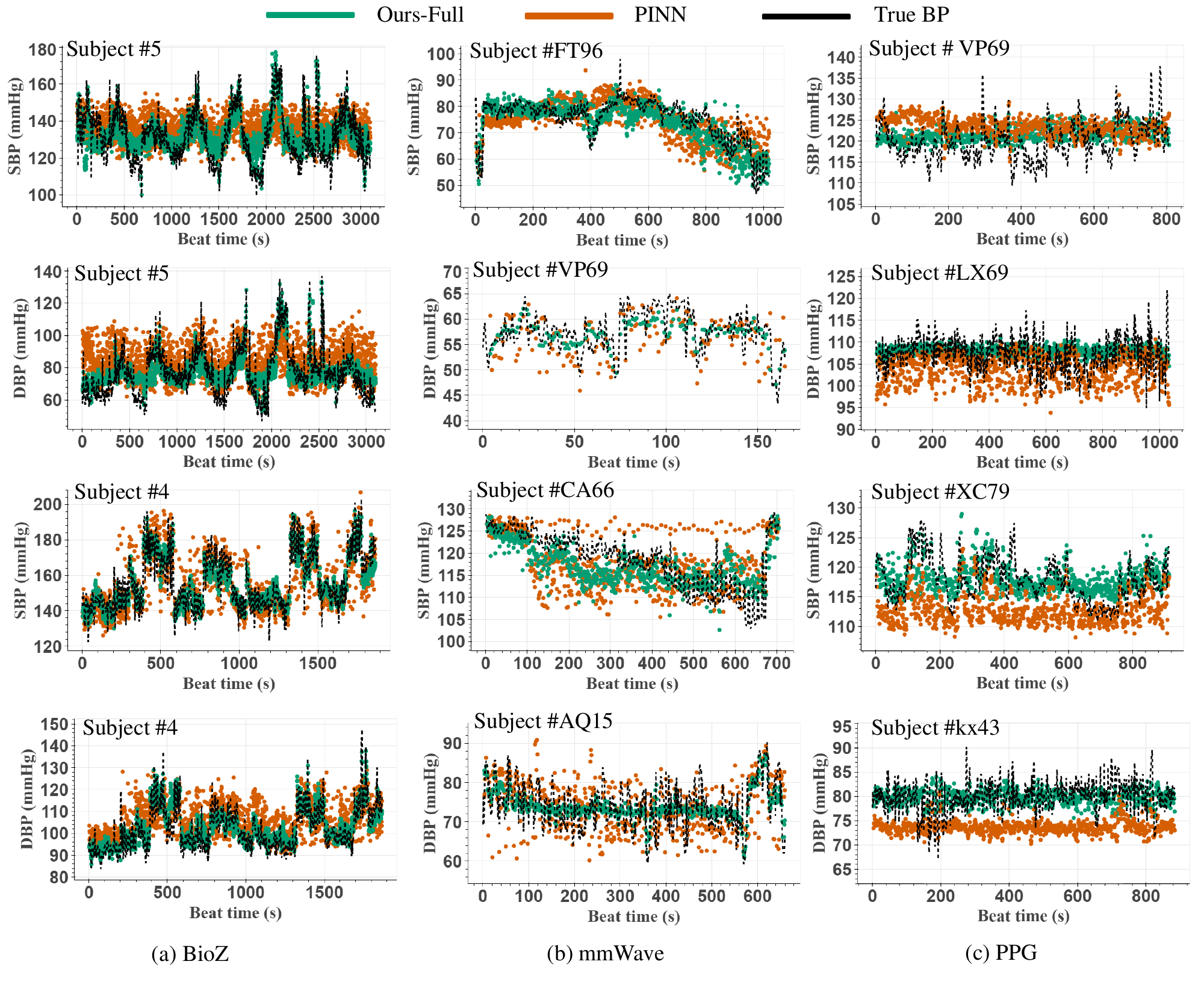}

	\caption{Beat-to-beat SBP and DBP estimation based on Ours-Full model (shown in green) and PINN (shown in orange) model trained with the same number of instances and corresponding true BP (shown in dashed black).
    The figures in each column correspond to the same signal, with the subject listed at the top of each figure. Ours-Full model shows a more precise fit to the reference BP. }
	\label{fig:beat-to-beat}

\end{figure*}

\begin{figure}[ht]
	\centering
	\includegraphics[width=0.5\textwidth]{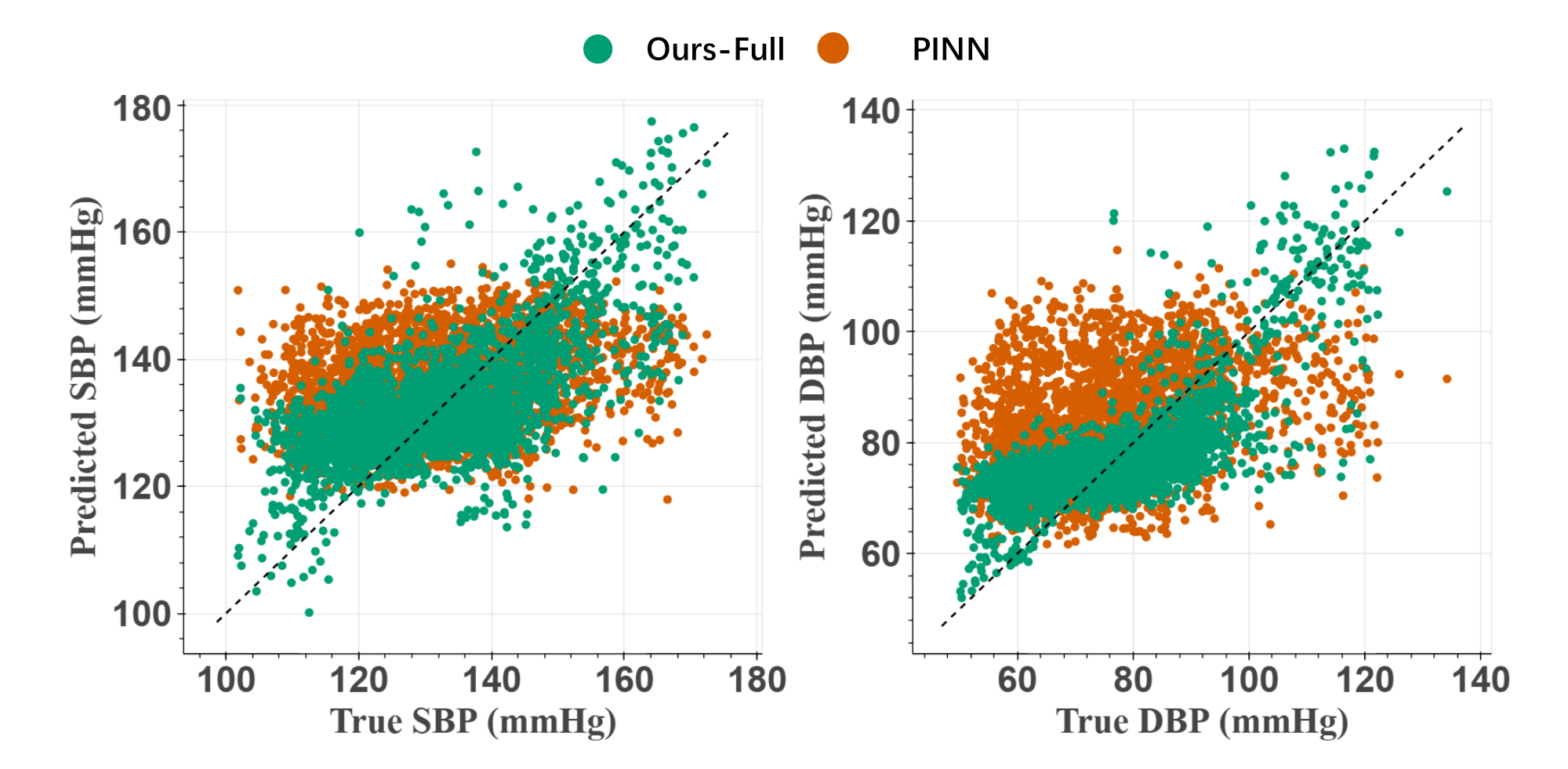} 
	\caption{Pearson’s correlation analysis with 2910 blood pressure values using subject \#5 in the Graphene-HGCPT dataset. Predictions of the Ours-Full model are in green, while predictions of the PINN model are in orange.} 
	\label{fig:correlations}
\end{figure}

Fig.~\ref{fig:beat-to-beat} visually compares the beat-to-beat SBP and DBP estimations using the Ours-Full model and PINN model. The results clearly demonstrate the effectiveness of the proposed model. Under identical training constraints, it is evident that the Ours-Full model outperforms the PINN model in capturing the personalized characteristics of subjects, yielding higher correlations and lower absolute errors. Ours-Full model consistently outperforms the PINN model across all three signal types (bioimpedance, mmWave, and PPG). Cardiovascular time series data typically involve intricate temporal patterns, where multiple variations (\emph{e.g.}, rising, falling, fluctuation) mix and overlap, making temporal variation modeling extremely challenging. Compared to the vanilla PINN, our PITN model is better equipped to capture such temporal information, resulting in a closer fit to the actual blood pressure values.

However, there are still a few failure cases, likely due to the limited number and inconsistent sampling observed points for certain subjects in the Blumio dataset. Additional training data could potentially improve the model's accuracy in BP estimation. Generally, the BioZ datasets (Ring-CPT and Graphene-HGCPT) contain more samples, leading to better model performance compared to the Blumio dataset.

As a proof-of-concept, Fig.~\ref{fig:correlations} presents Pearson's correlation visual analysis, by using subject \#5 from the Graphene-HGCPT as an example. 
We can clearly observe that the prediction points of the Ours-Full model are more closely aligned with the dashed line (where predicted values equal true values), while those of the PINN model are more dispersed. This further demonstrates the effectiveness of our proposed methods in personal modeling and in capturing discriminative variations and cardiovascular dynamics.

\subsection{Ablation Study}

In this subsection, we conduct ablation studies to examine the effectiveness of the proposed temporal block, adversarial training, and contrastive learning, respectively. 
\begin{table*}[!ht]
	\caption{Ablation study on proposed methods. \emph{Ours-Base}, \emph{Ours w/ adv}, and \emph{Ours-Full}, denote the baseline model, our model only with adversarial samples, and our full model, respectively. The performance improvement is compared to the PINN model. $\uparrow$ refers to the higher result being better and $\downarrow$ vice versa. The bold values represent the best performance and underlined values indicate the second best performance.}
	\centering
	\resizebox{0.9\textwidth}{!}{%
		\begin{tabular}{cccccccccccccccccc}
			\toprule
			{} & \multicolumn{3}{c}{{}} & \multicolumn{4}{c}{{Graphene-HGCPT}} & \multicolumn{4}{c}{{Ring-CPT}} &  \\
			\cmidrule(r){5-13}
			 {} & \multicolumn{3}{c}{\multirow{-2}{*}{{Method}}} & \multicolumn{2}{c}{{Corr $\uparrow$}} & \multicolumn{2}{c}{{RMSE $\downarrow$}} & \multicolumn{2}{c}{{Corr $\uparrow$}} & \multicolumn{2}{c}{RMSE $\downarrow$} \\
			\cmidrule(r){2-13}
			\multirow{-3}{*}{Model Name} & {Temporal block} & {Adversarial} &  {Contrastive} & {SBP} & {DBP} & {SBP} & {DBP} & {SBP} & {DBP} & {SBP} & {DBP} \\
			\midrule
                {PINN~\cite{sel2023physics}} & {\XSolidBrush} & {\XSolidBrush} &{\XSolidBrush} & {0.48} & {0.50} & {12.8} & {11.1} & {0.66} & {0.67} & {8.7} & {6.0}\\
			{Ours-Base} & {\Checkmark} & {\XSolidBrush}   & {\XSolidBrush} & {+0.12} & {+0.06} & {-1.4} & {-0.8} & {+0.08} & {+0.02} & {-1.7} & {-0.5} \\
			{Ours w/ adv } & {\Checkmark} &{\Checkmark}   & {\XSolidBrush} & \underline{+0.13} & \underline{+0.05} & \underline{-1.7} & \underline{-0.9} & \underline{+0.09} & \underline{+0.03} & \underline{-1.9} & \underline{-0.8}\\
			
			{\textbf{Ours-Full}} & {\Checkmark} & {\Checkmark} & {\Checkmark} & \textbf{+0.15} & \textbf{+0.07} & \textbf{-1.8} & \textbf{-1.0} & \textbf{+0.11} & \textbf{+0.05} & \textbf{-2.0} & \textbf{-0.8} \\
			
			\midrule
		\end{tabular}%
	}
	
	\label{tab: ablation study}

\end{table*}

\subsubsection{Temporal Block} 

As shown in Table~\ref{tab: ablation study}, with temporal modeling, the Ours-Base model outperforms the vanilla PINN model by a large margin. From the table, we can observe that our proposed temporal blocks have a significant influence on BP prediction, removing them will lead to a significant drop in performance. This demonstrates the vital significance of the proposed temporal block.  In terms of Pearson correlation, Ours-Base excels the vanilla PINN by relatively 25\% and 12\% in SBP and DBP estimation, while in terms of RMSE, Ours-Base outperforms the vanilla PINN by 11\% and 3\% relatively in SBP and DBP estimation.

\subsubsection{Adversarial Training} 

As shown in Table~\ref{tab: ablation study}, we can see augmenting the framework with adversarial samples (denoted as “Ours w/ adv”) enhances performance compared to the Ours-Base model, which utilizes only clean samples for training. This improvement suggests that minor perturbations in adversarial samples can enhance model performance, especially when training data is insufficient. Although these adversarial samples are designed to introduce perturbations through negative gradients, careful control of the generation steps and perturbations allows them to supplement the training data rather than disrupt it.

\begin{figure}[t]
	\centering
	\includegraphics[width=0.5\textwidth]{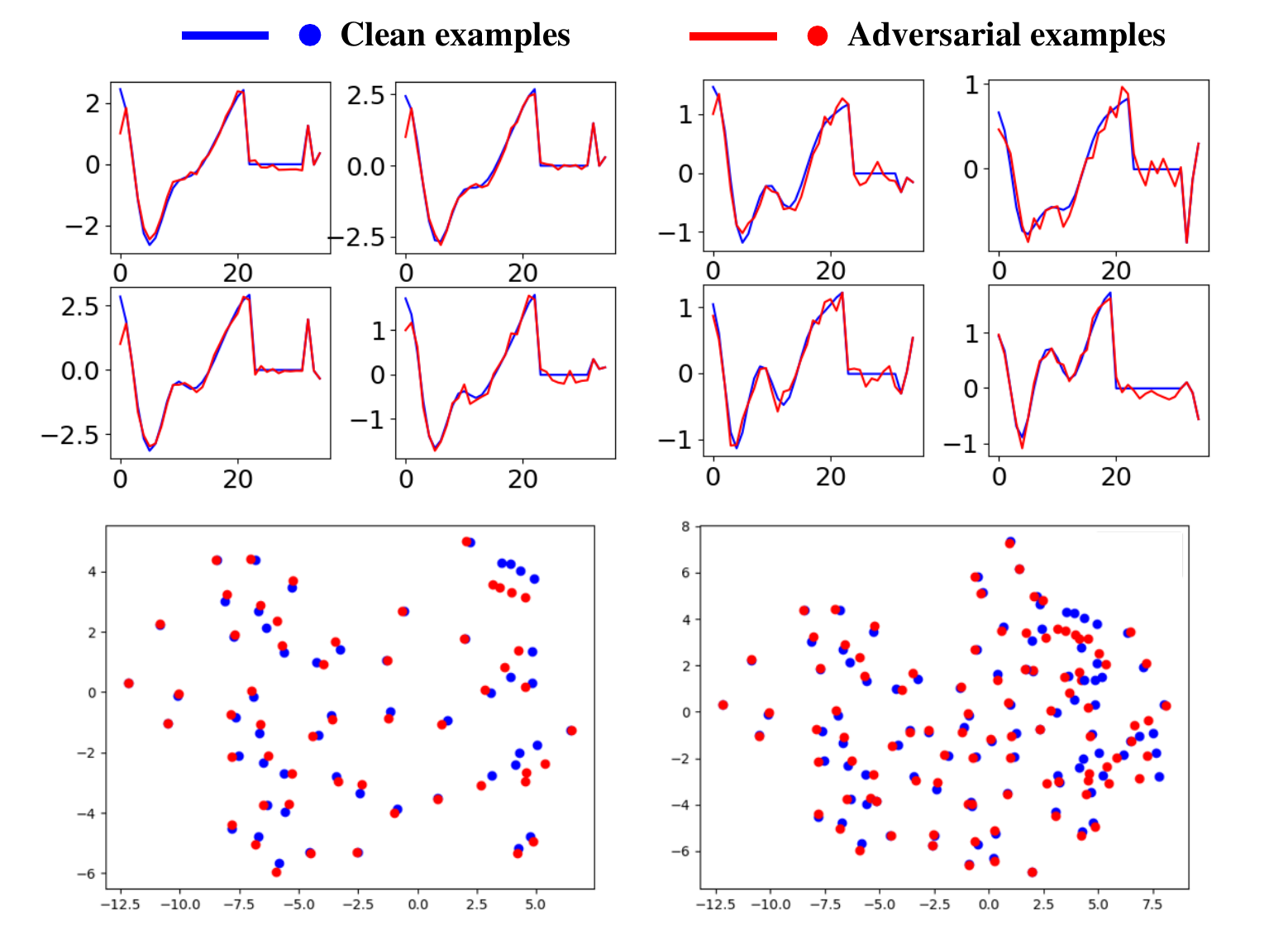}
	\caption{Visualizations on adversarial examples and clean examples in waveform (up) and T-SNE (down).}
	\label{fig: adversarial examples}
	
\end{figure}

It is evident from Fig.~\ref{fig: adversarial examples} that, although the generated adversarial examples exhibit differences in signal characteristics compared to the clean examples, their primary feature components remain closely aligned with the corresponding clean examples. Additionally, we compare various data augmentation methods in the supplementary material to further demonstrate the superiority of our approach.

However, as also shown in Table~\ref{tab: ablation study}, while adversarial examples improve the framework's performance on RMSE in two datasets, they may degrade Pearson’s correlation between the estimated and true BP values. To address this, we introduce contrastive learning to better capture BioZ signals with similar BP values, thereby improving correlation.

\subsubsection{Contrastive Learning} 

As shown in Table~\ref{tab: ablation study}, our full model achieves the best correlation between state-of-the-art methods and the Ours-Base model, demonstrating the effectiveness of incorporating contrastive learning to capture BP flow changes over time. As noted in~\cite{sel2023physics}, an increase in input BioZ data causes correlation to fluctuate. Thus our proposed contrastive learning approach addresses this issue by introducing an additional loss on bioimpedance signals that have relatively similar BP labels. As anticipated, by aligning bioimpedance signals with similar BP values, our framework can better capture BP dynamics, resulting in improved Pearson's correlation coefficients. Furthermore, by precisely modeling personal discriminative variations, our full model obtained slightly better performance in RMSE compared to the base model. Overall, it can be observed that, compared to the Ours-Base model, our full model effectively learns cardiovascular dynamics and personal discriminative variations.

\section{Conclusion}

In this paper, we presented an adversarial contrastive learning-based Physics-Informed Temporal Network for cuffless blood pressure estimation. Specifically, we introduced a temporal block within Physics-informed neural networks to extract intraperodic physiological features for temporal modeling. Additionally, we implemented adversarial training combined with contrastive learning to augment physiological time series data. Extensive experiments conducted on multimodal signals from the graphene-HGCPT, ring-CPT, and Blumio datasets, have demonstrated the superior effectiveness of our framework. In the future, we will apply the proposed framework to other medical applications, such as blood sugar monitoring.

\input{main.bbl}




\newpage






\end{document}